 \documentclass[final,5p,times,twocolumn,authoryear]{elsarticle}

\usepackage{amssymb}
\usepackage{lipsum}

\journal{Astronomy $\&$ Computing}

\begin{document}

\begin{frontmatter}



\title{Digitalization in Infrastructure Construction Projects: A PRISMA-Based Review of Benefits and Obstacles}


\author[first]{Mohammed Abdulsalam Alsofiani, Ph.D}
\affiliation[first]{organization={Formerly Texas A\&M University Instructor of Records},
           }

\begin{abstract}
The current study presents a comprehensive review of the benefits and barriers associated with the adoption of Building Information Modeling (BIM) in infrastructure projects, focusing on the period from 2013 to 2023. The research explores the manifold advantages offered by BIM, spanning the entire project life cycle, including planning, design, construction, maintenance, and sustainability. Notably, BIM enhances collaboration, facilitates real-time data-driven decision-making, and leads to substantial cost and time savings.

In parallel, a systematic literature review was conducted to identify and categorize the barriers hindering BIM adoption within the infrastructure industry. Eleven studies were selected for in-depth analysis, revealing a total of 74 obstacles. Through synthetic analysis and thematic clustering, seven primary impediments to BIM adoption were identified, encompassing challenges related to education/training, resistance to change, business value clarity, perceived cost, lack of standards and guidelines, lack of mandates, and lack of initiatives.

This review explores the benefits and barriers in the industry that are facing BIM adoption in infrastructure projects, giving an important perspective toward improving effective BIM adoption strategies, policies, and standards. Future directions for research and industry development are outlined, including efforts to enhance education and training, promote standardization, advocate for policy and mandates, and integrate BIM with emerging technologies. 

\end{abstract}



\begin{keyword}
BIM \sep Infrastructure \sep Benefits   \sep Obstacles \sep Adoption \sep Standard \sep PRISMA



\end{keyword}

\end{frontmatter}




\section{Introduction}
\label{introduction}

The construction industry faces persistent challenges related to project delays and cost overruns, which have significant implications for global infrastructure development. These delays are widely recognized as one of the most pressing issues in construction [1, 2]. They stem from various factors, including inadequate project design, poor scheduling, suboptimal project management, insufficient site supervision, design flaws, and communication breakdowns [3, 4].
The problem is not confined to specific regions or economic classifications. For instance, in China, rework is a prevalent issue, and its root causes remain unclear, posing a substantial challenge to construction businesses [5]. In multiple countries, public infrastructure projects often exceed budgeted costs and miss projected completion deadlines, raising concerns about the effectiveness of project management [6, 7]. Even regions with substantial infrastructure investments, such as Qatar, have experienced significant cost overruns and time delays in projects [8]. These persistent issues cast doubt on the construction industry's ability to meet project deadlines and stay within budget constraints without compromising quality [9].
However, a potential solution lies in Building Information Modeling (BIM), a well-established approach in architecture and construction. BIM allows for the creation and management of digital representations of physical and functional aspects of buildings and infrastructure, offering real-time access to critical project information for all stakeholders, thereby streamlining project execution and lifecycle management [10, 11].
The efficient use of infrastructure, which often requires large investment, is projected, in the long run, to provide major advantages to society [12]. For example, one of the National Priorities for the UK is the vision for infrastructure improvement and development by implementing and adopting digital skills [13]. Therefore, this article review tries to answer critical questions related to BIM adoption in infrastructure projects which are: 

\subsection{What are the benefits of BIM adoption in infrastructure projects?}
Understanding the benefits of BIM adoption in infrastructure projects is crucial for stakeholders, as it allows them to recognize the potential advantages and improvements that can be gained from implementing BIM technology. By identifying these benefits, organizations can make informed decisions regarding BIM adoption, potentially leading to increased project efficiency, cost savings, and improved project outcomes.
\subsection{What are the barriers to BIM adoption in infrastructure projects?}
Identifying the barriers to BIM adoption in infrastructure projects is essential for stakeholders as it helps them pinpoint the challenges and obstacles that may impede the successful implementation of BIM technology. Recognizing these barriers enables organizations to develop practical strategies and solutions to overcome these obstacles, ultimately facilitating the adoption of BIM and maximizing its potential benefits.
By addressing these questions, stakeholders can gain valuable insights into both the advantages and challenges associated with BIM adoption, empowering them to make informed decisions and implement effective approaches to successfully integrate BIM into their infrastructure projects.

\section{Methodology}

To address the central question of identifying the barriers to adopting Building Information Modeling (BIM) for infrastructure projects, we first recognize the need to establish the benefits of BIM technologies for such projects. Therefore, our review comprises two key stages.
\subsection{Stage One: Identifying BIM Benefits for Infrastructure Projects}
In the initial stage, our objective is to identify the advantages of implementing BIM in infrastructure projects. To achieve this, we employed data visualization tools to create a visual representation of the relationships between the articles under consideration. Specifically, we used VOSviewer, a software tool designed for visualizing indexed academic publications [14]. VOSviewer allows us to illustrate co-occurrence patterns and relationships among keywords and authors within a set of articles [15]. In this visualization, the color of the lines reflects the strength of the connections between keywords.
\subsection{Stage Two: Systematic Review}
Subsequently, we conducted a content analysis, chosen for its accessibility and systematic, replicable nature in extracting information from publications [16]. This methodological choice is guided by the need to analyze the textual content of selected articles [17]. Our systematic literature review adhered to the Preferred Reporting Items for Systematic Reviews and Meta-Analyses (PRISMA) statement, which offers a framework for evidence-based reporting. The objective of this qualitative approach is to develop a comprehensive understanding of the primary barriers associated with the adoption of BIM in infrastructure projects.

\subsubsection{PRISMA Protocol}
Our systematic review followed the three primary stages outlined in the PRISMA protocol: planning, review, and reporting. The planning stage encompassed setting review objectives, establishing databases for the study, defining keywords, and outlining inclusion and exclusion criteria. These steps were crucial in identifying impediments to the adoption of BIM in infrastructure projects. The review stage involved a qualitative synthesis to identify articles directly relevant to our specific research aim. Lastly, the reporting phase comprised a descriptive analysis and the categorization of identified barriers to BIM adoption in infrastructure projects.
To validate the qualitative data, we employed a PRISMA flow chart. This flow chart serves as a validation tool for our content analysis, illustrating the progression from initial article selection based on inclusion criteria, through the exclusion process, to the final selection of articles [18].
\section{BIM Benefits for Infrastructure Projects}
To explore the benefits of BIM in the scientific domain, a wide literature review was performed with a specific date range filtration from 2013 to 2023 using Web of Science as a main database to find relative articles that address the potential benefits of BIM in infrastructure projects. The keywords “Building Information Modeling infrastructure” were searched in the abstract section. As a result, 447 papers were identified and exported into RIS format to be uploaded into a data visualization tool to visualize the relationship between these articles by finding the occurrence of the keywords. VOSviewr, a software for visualizing indexed academic publications was used [14]. This tool is designed to illustrate the co-occurrence aggregates and relationship between keywords and the author of a set of articles [15]. The color of the line, on the other hand, reflects the connectivity of the keywords. 
Figure 1 shows that the word “benefits” was one of the revealed recurring keywords and it has a strong connection with other keywords such as cost, performance, sustainability, prediction, behavior, and life cycle. Therefore, a different approach for conducting a literature review for BIM benefits in infrastructure was also conducted. Additionally, the map of the keyword occurrence over time, Figure 1, shows the connection between the word benefits and new technologies such as Digital Twin (DT), Internet of Things (IoT), and blockchain. This reflects the importance of BIM adoption for the future implementation of Construction 4.0 technologies.
\begin{figure}
	\centering 
	\includegraphics[width=0.4\textwidth, angle=-0]{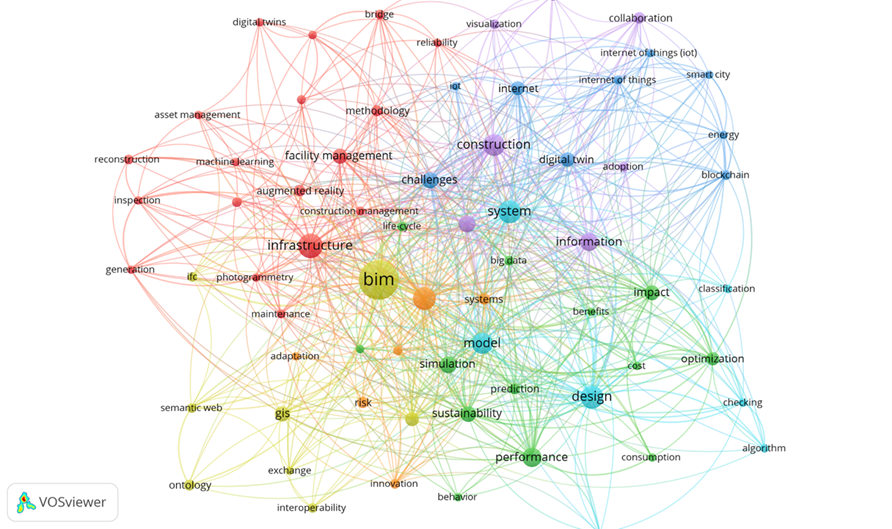}	
	\caption{The Occurrence and Relationships Among Keywords in 447 BIM-Related Papers} 
	\label{fig_mom0}%
\end{figure}
Various approaches, both theoretical and practical, have been suggested demonstrating the benefits of BIM for infrastructure projects. A systematic review approach, conducted by Bradley et al. [19] and Doumbouya et al. [20], identified the benefits BIM in infrastructure and revealed promising advantages. Evidently, BIM offers a verity of benefits to the construction industry. The identified benefits of BIM in infrastructure projects include more effective collaboration and communication between stakeholders, automation of recurring tasks, and better utilization and analysis of the project information [19]. 
\subsection{Theoretical BIM Benefits for Infrastructure}
In their systematic review of BIM in transportation infrastructure, Costin et al. [21] conclude that BIM technologies can help the AEC industry regarding infrastructure projects in nine aspects of the life cycle: planning, design, scheduling, construction, maintenance, structural health monitoring, as-built data documentation, renovation, and behavior modeling and prediction. 
For planning, BIM is a powerful and easy tool for life cycle analysis and strategic planning [22]. Especially, applying BIM 4D (time) and more advanced BIM 5D [21] to massive infrastructure projects may have a bigger impact on the asset values; the more complex and larger the project is, the larger the benefits earned [23]. 
When it comes to design, BIM brings the best visual experience for designing infrastructure projects, which improves collaboration and coordination among designers and engineers [24]. In addition to considerable advances in the design process, fewer omissions and mistakes, as well as reductions in disputes on the construction site, are some of the most valued advantages of adoption from the design stage forward [25]. In addition, BIM can help reduce the energy consumption of a construction project by providing effective tools via BIM for design optimization, which leads to waste reduction [26]. Although BIM can help in energy consumption analysis, several tools and plug-ins were developed by researchers and third-party organizations to address the goal of energy analysis and carbon emission calculation for better sustainable design solutions. For example, Jalaei et al. [27] developed a plug-in application that assesses the design based on the LEED green building rating system, which as a result reduces the embodied carbon compared to the traditional design process.   
For project scheduling, the real-time visual scheduling capabilities provided by BIM make it simpler to identify activities that are operating simultaneously and might cause conflicts during the construction phase. This can also help in traffic management during massive infrastructure development [28]. Also, in construction, BIM applications have the capabilities to effectively manage on-site activities and significantly reduce the number of requests for information (RFI) and change orders [29]. Furthermore, BIM applications can assist crews on-site to understand complex components of the design and give them a better understanding of how to perform specific construction activities [30]. 
When it comes to the maintenance and conservation of infrastructure assets, BIM plays a major role as a source of data that has been collected during the lifecycle of the project [21]. This data can be used to maintain an ideal operational status, predict the needs for repairs, and enhance the safety of the infrastructure environment [31-34]. In addition, one of the most promising benefits that BIM systems may provide is the ability to simulate human behavior. BIM applications may assess and anticipate behavior modeling and forecasting models based on data from the original BIM model, inspection visits, and sensors, among other sources [21]. 
It is important to highlight the positive impact of BIM as a significant carbon emission reduction factor in the AEC industry. The scientific research studies that provide insights and evidence of such benefits are increasing over time. For example, Hao et al. [35] conducted a study to assess the effectiveness of using BIM technologies to measure carbon emissions of new construction infrastructure projects and compared the results with conventional construction methods. BIM was found to be an effective and efficient tool for measuring carbon emissions of new construction projects, which opens the door for alternative materials, different methods, and better energy consumption plans.  
In a recent study by Zhao et al. [36], an existing building was digitally imported to BIM via laser scanning to evaluate retrofitting schemes based on energy efficiency to reduce energy consumption. According to this case study, the reduction of energy consumption has reduced carbon emissions by 85.9 percent, which shows a significant impact of BIM technologies.

\subsection{Practical Evidence of BIM Benefits for Infrastructure}
Exploring BIM benefits for infrastructure projects through case studies gives a practical perception of BIM advantages. For instance, Vignali et al. [37] used BIM tools for upgrading a road project in northern Italy. They found that BIM empowers the optimization and validation of the project at the design phase, and it superbly illustrates the project in a real 3D environmental context. 
	Onungwa et al. [38] explored the benefits that BIM can bring to the infrastructure of a smart city. They created a BIM model for the master plan of an existing smart city in the Atlanta Campus of the Georgia Institute of Technology. As a result of this study, they concluded that the BIM benefits for infrastructure include better communication between stakeholders, real-time activity tracking, and effective data visualization.
Literature shows that the BIM model can help with cost and time reduction for infrastructure projects [39, 40].  For example, Fanning et al. [29] studied the impact of BIM implementation in a bridge construction project in Colorado. They compared two similar ongoing bridge construction projects, employing BIM in one of them. The findings of this study showed that the bridge construction project with BIM implementation may have reduced the project cost by 5-9 percent by reducing the change orders and rework. Additionally, the implementation of BIM in a bridge construction project may have a positive impact on infeasible complex projects and reduce requests for information and change orders [29]. 
In a recent study, Shin et al. [41] compared six subway extension design projects in South Korea completed by six different design firms where three firms were BIM-based design firms and the other three were non-BIM, traditional design process firms. This study found that there was a 2.9 percent increase in productivity with BIM-based design, with a decrease in their man-hours, which resulted in 103.5 fewer days than the non-BIM project on average and required less professional labor force to complete the project design. Another study conducted by Heikkilä et al. [42] evaluated BIM benefits for three infrastructure projects from 2012 to 2022 during the design and construction phases. This study found clear economic and technical benefits such as a reduction in construction and design time, emphasized machine control, better build quality, and better quality control, which resulted in saving to up 7 percent of design and construction costs.   
Bensalah et al. [43] performed a full BIM integration in a railway project in Morocco to examine the benefits to the industry. The findings confirmed that the integration of BIM into railway projects can provide several advantages; cost control, rework avoidance, improved detection of interface issues, improved project scheduling, reduced construction phase duration, and demonstrated facility management for maintenance and preservation. Moreover, efficient planning and earthwork productivity were a result of developing a BIM model for the renovation of an approximate 1-mile sewerage network in South Korea, and it serves as a data storage to be used in the future for operating and management peruses [44]. For example, in Wuhan (China), a BIM model was developed to help assess risk associated with a metro tunnel project, which shows an effective approach to identifying a safety hazard regarding water gushing during construction [45]. 
The use of a BIM model in a construction project allows the ability to obtain more valid and accurate real-time data for built asset operation and management throughout sensor implementation [46-49]. For example, research shows that the data produced by an implemented BIM model of sensors in    a highway in the UK allowed for efficient management and operation practices [50]. Moreover, an airport runway in Madrid was developed in a BIM model that was updated frequently by new real-time data collected during a regular routine inspection [51]. This led to less manual work, sustainable control of the built asset, and efficient way to store and update the facility data. 
\section{{Barriers to BIM adoption in infrastructure projects}
}
It is important to present the results of the PRISMA protocol systematic review in the context of describing the review procedure because each step has results that represent part of the final holistic result. 
\subsection{Stage One}
Planning the literature review procedure, which included the review objectives, database, keywords, and inclusion and exclusion criteria. This stage aimed to come up with a set of impediments to adopting BIM in infrastructure projects. 
The chosen databases for conducting this systematic review were Web of Science and Science Direct. As Web of Science provides multidisciplinary sources from about 8,500 of the most well-known high-impact research journals worldwide. Additionally, the reason behind choosing Science Direct as a second database is that it is a subsidiary of Elsevier and contains over 2000 Elsevier-published journals.  
The following keywords “Building Information Modeling”, “adoption”, and “infrastructure” were searched in the title and abstract field in both databases. The keyword search was conducted in March 2023; A total of 175 results were obtained with 149 results from Web of Science and 26 results from Science Direct as shown in Figure 2. 
\begin{figure}
	\centering 
	\includegraphics[width=0.4\textwidth, angle=-0]{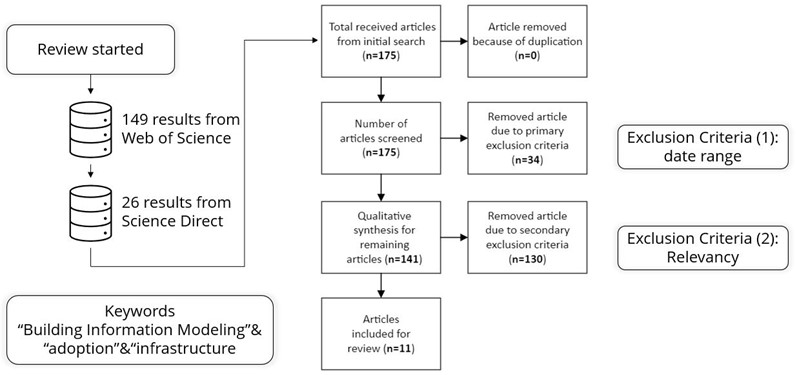}	
	\caption{Systematic Review Flowchart Procedure} 
	\label{fig_mom01}%
\end{figure}

The inclusionary criteria were applied, which are peer-reviewed Journal articles and published in English. Meanwhile, the exclusionary criteria were duplicated records, preceding papers, books and chapters, and reports. To reduce the number of results, a date range was chosen from 2015 to 2023. This led to a total of 141 journal articles that needed to be reviewed.
\subsection{Stage Two}
141 journal articles were reviewed, using secondary criteria involving qualitative synthesis. The process aimed to identify articles relevant to BIM adoption barriers in infrastructure projects. The review of the 141 journal articles passed through secondary criteria, which is a qualitative synthesis to identify relevant articles to the specific aim which is the barriers to BIM adoption in infrastructure projects. As a result, 11 journal articles identified barriers to BIM adoption in infrastructure projects as shown in Table 1.   
\begin{table}
    \centering
    \begin{tabular}{|l|l|}
        \hline
        Database & Authors \\ \hline
        Science Direct & Love et al., 2015 \\ \hline
        Web of Science & Vass \& Karrbom Gustavsson, 2017 \\ \hline
        Science Direct & Heaton et al., 2019 \\ \hline
        Web of Science & Hong et al., 2019 \\ \hline
        Web of Science & Marefat et al., 2019 \\ \hline
        Web of Science & Roy et al., 2020 \\ \hline
        Web of Science & Umar, 2021 \\ \hline
        Web of Science & Arrotéia et al., 2021 \\ \hline
        Web of Science & Evans \& Farrell, 2020 \\ \hline
        Web of Science & Alemayehu et al., 2021 \\ \hline
        Science Direct & Liu et al., 2019 \\ \hline
    \end{tabular}
    \caption{The Identified Journal Articles Characteristic (Database/Author)}
    \label{tab:my_label}
\end{table}
\subsection{Stage Three}

To report the findings, a descriptive analysis was conducted for 11 journal articles to draw the barriers to BIM adoption in infrastructure projects. Hence, a total of 74 barriers to BIM adoption in infrastructure were identified.
The number of identified barriers was quite large, so reducing this number was necessary. To achieve this, duplicated or repeated barriers were excluded. Furthermore, a labeling technique based on meaning and definitions was employed, which resulted in 7 main labels indicating the primary barriers with a specified definition. These primary barriers were change resistance, lack of education and training, lack of awareness about BIM, perceived cost of BIM adoption, lack of standards and guidelines for BIM implementation, lack of BIM mandatory, and lack of BIM initiatives. 
Nevertheless, with all the benefits that were discussed in the literature review chapter, the adoption and implementation of BIM in infrastructure projects encountered many challenges. This systematic review identified 72 barriers that inhibit the adoption of BIM in infrastructure. One of the important barriers is the lack of standardization and guidelines that regulate the relationship between all beneficial parties of BIM such as responsibility and ownership. Another important barrier to BIM adoption in infrastructure is the lack of training and education. Moreover, the lack of initiative in terms of technical support and motivation was repeatedly reported. Lastly, the lack of client demand and lack of BIM mandatory is one of the most mentioned barriers. However, all 72 barriers were clustered based on the 7 label criteria. 
The first label, Change Resistance, was defined in the literature in so many ways that all led to a similar meaning. For example, Oreg [52], defined change resistance as individuals who are not willing to adopt new ideas. The second label was Education and Training, which is a common barrier to BIM adoption in the literature. It is necessary to set a systematic learning process, training, and education, in order to adopt and acquire new knowledge [53]. The third label, Unclear Business Value of BIM, is also a common barrier in the AEC industry. The business value of BIM includes BIM benefits and the prerequisite elements, such as awareness of the values that BIM can bring [54]. Awareness is defined as the individual ability to link between current knowledge and new knowledge and apply that information to different or new situations [55-57]. The fourth label was the cost of adoption.  While the cost of adoption is a burden to get the benefit of technology adoption, people have a perceived cost to it. Indeed, the time and effort to adopt new technology is considered a perceived cost [58]. The fifth label was standards and guidelines, which are a series of documents produced by a technical committee that indicates activity procedures and requirements to assist practitioners and stakeholders [59]. The sixth label was Government-Mandates, which are regulations that firms are forced to comply with under the supervision of the government or a societal organization [60]. The seventh label was Lack of Initiative. The initiative aims to formalize and motivate stakeholders by providing several types of support such as technical support [61]. Technical support, on the other hand, is defined as the experience and knowledge of software and hardware performed by skilled individuals [62, 63].
 \section{Conclusion }
 This review article highlights the needs, benefits, and challenges of BIM adoption in infrastructure projects, offering crucial insights for enhancing project efficiencies, recommending policies and standards, and guiding future research to support the construction industry's advancement.
The extensive review of literature from 2013 to 2023 highlights the manifold benefits of adopting Building Information Modeling (BIM) in infrastructure projects. BIM's advantages span the entire project life cycle, from planning and design to construction, maintenance, and sustainability. Notably, BIM fosters enhanced collaboration, real-time data-driven decision-making, and substantial cost and time savings.
Theoretical and practical evidence demonstrates that BIM optimizes project management, reduces errors, improves communication, and aids in sustainable design and energy consumption analysis. Case studies underscore BIM's real-world impact, with significant reductions in project costs and construction durations observed. Moreover, BIM's integration with emerging technologies like Digital Twin and the Internet of Things positions it as a cornerstone for Construction 4.0. However, the adoption of BIM in infrastructure projects not only streamlines processes but also paves the way for more efficient, sustainable, and cost-effective construction practices. The findings underscore the pivotal role of BIM in shaping the future of infrastructure development.
Additionally, a systematic literature review was conducted to ascertain the barriers obstructing the adoption of BIM within the infrastructure industry. The systematic review adhered to the PRISMA protocol, leading to the selection of 11 studies for in-depth analysis. The analysis revealed a total of 74 obstacles hindering the adoption of BIM in infrastructure projects. Since many of these obstacles were found in multiple studies, redundant obstacles were eliminated, and a synthetic analysis was conducted to categorize the remaining challenges based on thematic clusters. The procedure used was inductive coding, in which the code was generated from the obtained data. As a result, seven main obstacles were identified as primary impediments to BIM adoption in infrastructure projects, as shown in Figure 3.
\begin{figure}
	\centering 
	\includegraphics[width=0.4\textwidth, angle=-0]{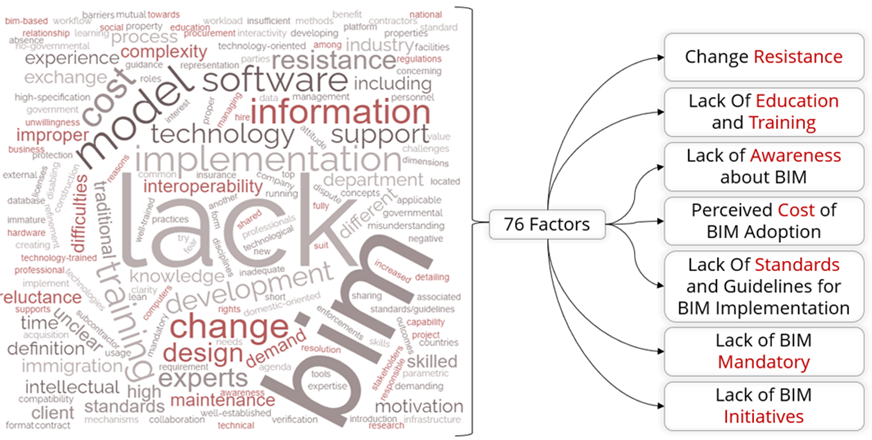}	
	\caption{Results of the Inductive Analysis and Clustering Procedure for BIM Barriers in Infrastructure Projects} 
	\label{fig_mom02}%
\end{figure}

These obstacles are lack of education/training, resistance to change, unclear business value, perceived cost, lack of standards and guidelines, lack of mandates, and lack of initiatives. 
\section{Contribution}
Specific contributions to understanding and implementing Building Information Modeling in infrastructure projects: First, the paper provides a detailed review of the benefits and barriers associated with the adoption of BIM over the last decade (2013–2023). Second, it accentuates how BIM enhances collaboration, helps in making real-time data-driven decisions, and accounts for tremendous cost and time savings during the life cycle of the project (from planning, design, construction, and maintenance to sustainability). Second, the research identifies and classifies 74 barriers to BIM adoption and synthesizes them into seven main barriers: (1) education and training challenges, (2) resistance to change, (3) lack of clarity of business value, (4) perceived to be costly, (5) lack of standards, guidelines, (6) mandates, and initiatives. A systematic literature review of this paper follows the PRISMA protocol as adopted to provide a robust methodological framework for further research on BIM adoption barriers.
Finally, future research and industry development are portrayed. This makes it evident how future research should focus more on education and training, advancement in standardization, lobbying for policy and mandates, integration of BIM with emerging technologies, and more longitudinal studies that could embrace the learning of long-term impacts. Moreover, practical proof has been given with case studies showing a significant reduction in project cost and duration due to the utilization of BIM, thus practically validating the benefits of BIM to the infrastructure. Thematic clustering of literature findings is presented, therefore giving a structured understanding toward the benefits and barriers to BIM adoption. Discussion of the integration of BIM with technologies such as Digital Twin and the Internet of Things hence positions BIM as a stepping stone toward the future of Construction 4.0. The following contributions, therefore, provide a deep insight roadmap to the stakeholders concerned with the construction business to comprehend and explore the complexities relating to BIM adoption in infrastructure-related studies.
\section{Future Directions}
The extensive review of literature from 2013 to 2023 has illuminated the manifold benefits of adopting Building Information Modeling (BIM) in infrastructure projects, highlighting its transformative potential across the project life cycle. To chart a path forward, several key directions emerge. First and foremost, addressing the seven primary barriers identified in this study, including education/training, resistance to change, and business value clarity, should be a top priority. Future research can delve deeper into strategies and frameworks to mitigate these challenges and facilitate widespread BIM adoption. Ensuring a BIM-ready workforce remains crucial, necessitating ongoing efforts to refine education and training programs within both academic institutions and industry settings. Additionally, the development and implementation of standardized practices and guidelines can contribute to smoother BIM adoption, warranting further exploration. Encouraging governments and regulatory bodies to establish mandates and initiatives supporting BIM adoption is vital, and future research can investigate the impact of policy measures in various regions and industries. As BIM converges with emerging technologies like Digital Twin and the Internet of Things, further exploration is needed to fully harness their synergies. Practical applications, interoperability challenges, and holistic approaches to Construction 4.0 should be explored. Finally, longitudinal studies tracking the long-term impact of BIM on infrastructure projects can offer invaluable insights into performance, cost savings, and sustainability benefits over extended periods. In conclusion, this research not only underscores the transformative potential of BIM but also provides a roadmap for future research aimed at addressing barriers, refining practices, and driving the continued evolution of BIM as a cornerstone of the construction industry.

\section{Acknowledgments}

This research is part of my doctoral dissertation conducted at Texas A\&M University. I would like to thank my advisors, Dr. Stephen Caffey, Dr. Edelmiro Escamilla, Dr. Michael Lewis, and Dr. Kim Dooley, for their guidance and support throughout this project. Additionally, I would like to express my deep appreciation to my country, Saudi Arabia, for funding and supporting me during this journey.

\section{Refrences}
1.	Gardezi, S.S.S., I.A. Manarvi, and S.J.S. Gardezi, Time extension factors in construction industry of Pakistan. Procedia Engineering, 2014. 77: p. 196-204.

2.	Ismail, A., Time and Cost overrun in public construction projects in Qatar. Master’s project, Qatar University, Doha, Qatar, 2014.

3.	Marzouk, M.M. and T.I. El-Rasas, Analyzing delay causes in Egyptian construction projects. J Adv Res, 2014. 5(1): p. 49-55.

4.	Zidane, Y.J.-T. and B. Andersen, The top 10 universal delay factors in construction projects. International Journal of Managing Projects in Business, 2018.

5.	Ye, G., et al., Analyzing causes for reworks in construction projects in China. Journal of Management in Engineering, 2015. 31(6): p. 04014097.

6.	França, A. and A.N. Haddad, Causes of construction projects cost overrun in Brazil. International Journal of Sustainable Construction Engineering and Technology, 2018. 9(1): p. 69-83.

7.	Santos, H.d.P., C.M.D. Starling, and P.R.P. Andery, Um estudo sobre as causas de aumentos de custos e de prazos em obras de edificações públicas municipais. Ambiente Construído, 2015. 15: p. 225-242.

8.	Senouci, A., A. Ismail, and N. Eldin, Time delay and cost overrun in Qatari public construction projects. Procedia engineering, 2016. 164: p. 368-375.

9.	Johnson, R.M. and R.I.I. Babu, Time and cost overruns in the UAE construction industry: a critical analysis. International Journal of Construction Management, 2020. 20(5): p. 402-411.

10.	Blanco, F.G.B. and H. Chen, The implementation of building information modelling in the United Kingdom by the transport industry. Procedia-Social and Behavioral Sciences, 2014. 138: p. 510-520.

11.	Georgiadou, M.C., An overview of benefits and challenges of building information modelling (BIM) adoption in UK residential projects. Construction Innovation, 2019.

12.	Damidavičius, J., M. Burinskienė, and R. Ušpalytė-Vitkūnienė, A monitoring system for sustainable urban mobility plans. Baltic journal of road and bridge engineering, 2019. 14(2): p. 158-177.

13.	Business, D.f., Industrial Strategy: building a Britain fit for the future. 2017.

14.	Van Eck, N.J. and L. Waltman, Manual for VOSviewer version 1.6. 8. CWTS meaningful metrics. Universiteit Leiden, 2018.

15.	van Eck, N.J. and L. Waltman, Citation-based clustering of publications using CitNetExplorer and VOSviewer. Scientometrics, 2017. 111(2): p. 1053-1070.

16.	Bush, A.A., M. Amechi, and A. Persky, An Exploration of Pharmacy Education Researchers' Perceptions and Experiences Conducting Qualitative Research. Am J Pharm Educ, 2020. 84(3): p. ajpe7129.

17.	Stemler, S., An overview of content analysis. Practical assessment, research, and evaluation, 2000. 7(1): p. 17.

18.	Pati, D. and L.N. Lorusso, How to write a systematic review of the literature. HERD: Health Environments Research \& Design Journal, 2018. 11(1): p. 15-30.

19.	Bradley, A., et al., BIM for infrastructure: An overall review and constructor perspective. Automation in Construction, 2016. 71: p. 139-152.

20.	Doumbouya, L., G. Gao, and C. Guan, Adoption of the Building Information Modeling (BIM) for construction project effectiveness: The review of BIM benefits. American Journal of Civil Engineering and Architecture, 2016. 4(3): p. 74-79.

21.	Costin, A., et al., Building Information Modeling (BIM) for transportation infrastructure–Literature review, applications, challenges, and recommendations. Automation in construction, 2018. 94: p. 257-281.

22.	Shin, H., et al., Analysis and design of reinforced concrete bridge column based on BIM. Procedia Engineering, 2011. 14: p. 2160-2163.

23.	Jones, S. and D. Laquidara-Carr, The business value of BIM for infrastructure, SmartMarket report, Rep. to Dodge Data \& Analytics, Balford, MA, 2017. 2017.

24.	Shim, C., N. Yun, and H. Song, Application of 3D bridge information modeling to design and construction of bridges. Procedia Engineering, 2011. 14: p. 95-99.

25.	Liu, S., et al., The driving force of government in promoting BIM implementation. J. Mgmt. \& Sustainability, 2015. 5: p. 157.

26.	Haruna, A., N. Shafiq, and O. Montasir, Building information modelling application for developing sustainable building (Multi criteria decision making approach). Ain Shams Engineering Journal, 2021. 12(1): p. 293-302.

27.	Jalaei, F., F. Jalaei, and S. Mohammadi, An integrated BIM-LEED application to automate sustainable design assessment framework at the conceptual stage of building projects. Sustainable Cities and Society, 2020. 53: p. 101979.

28.	Liapi, K.A. 4D visualization of highway construction projects. in Proceedings on Seventh International Conference on Information Visualization, 2003. IV 2003. 2003. IEEE.

29.	Fanning, B., et al., Implementing BIM on infrastructure: Comparison of two bridge construction projects. Practice periodical on structural design and construction, 2015. 20(4): p. 04014044.

30.	Matsumura, R., et al. CIM initiatives for Tokyo Gaikan Expressway Tajiri construction project. in 16th Int Conf Comput Civ Build Eng ICCCBE2016) p. 2016.

31.	Abudayyeh, O. and H.T. Al-Battaineh, As-built information model for bridge maintenance. Journal of computing in civil engineering, 2003. 17(2): p. 105-112.

32.	DiBernardo, S. Integrated modeling systems for bridge asset management—Case study. in Structures Congress 2012. 2012.

33.	Marzouk, M., et al. On the use of building information modeling in infrastructure bridges. in Proceedings of the 27th International Conference on Applications of IT in the AEC Industry, Cairo, Egypt. 2010.

34.	Yin, Z.-h., et al., Integration research and design of the bridge maintenance management system. Procedia Engineering, 2011. 15: p. 5429-5434.

35.	Hao, J.L., et al., Carbon emission reduction in prefabrication construction during materialization stage: A BIM-based life-cycle assessment approach. Sci Total Environ, 2020. 723: p. 137870.

36.	Zhao, L., et al., Digital-Twin-based evaluation of nearly zero-energy building for existing buildings based on scan-to-BIM. Advances in Civil Engineering, 2021. 2021: p. 1-11.

37.	Vignali, V., et al., Building information Modelling (BIM) application for an existing road infrastructure. Automation in Construction, 2021. 128: p. 103752.

38.	Onungwa, I., N. Olugu-Uduma, and D.R. Shelden, Cloud BIM Technology as a Means of Collaboration and Project Integration in Smart Cities. SAGE Open, 2021. 11(3): p. 21582440211033250.

39.	Ninić, J., et al., Computationally efficient simulation in urban mechanized tunneling based on multilevel BIM models. Journal of Computing in Civil Engineering, 2019. 33(3): p. 04019007.

40.	Osello, A., N. Rapetti, and F. Semeraro. BIM methodology approach to infrastructure design: Case study of Paniga tunnel. in IOP Conference Series: Materials Science and Engineering. 2017. IOP Publishing.

41.	Shin, M.-H., J.-H. Jung, and H.-Y. Kim, Quantitative and Qualitative Analysis of Applying Building Information Modeling (BIM) for Infrastructure Design Process. Buildings, 2022. 12(9): p. 1476.

42.	Heikkilä, R., T. Kolli, and T. Rauhala. Benefits of Open InfraBIM–Finland Experience. in ISARC. Proceedings of the International Symposium on Automation and Robotics in Construction. 2022. IAARC Publications.

43.	Bensalah, M., A. Elouadi, and H. Mharzi, Overview: the opportunity of BIM in railway. Smart and Sustainable Built Environment, 2019.

44.	Sharafat, A., et al., BIM-GIS-based integrated framework for underground utility management system for earthwork operations. Applied Sciences, 2021. 11(12): p. 5721.

45.	Zhang, L., et al., Bim-based risk identification system in tunnel construction. Journal of Civil Engineering and Management, 2016. 22(4): p. 529-539.

46.	Hu, C. and S. Zhang. Study on BIM technology application in the whole life cycle of the utility tunnel. in International Symposium for Intelligent Transportation and Smart City (ITASC) 2019 Proceedings: Branch of ISADS (The International Symposium on Autonomous Decentralized Systems). 2019. Springer.

47.	Huang, P., et al. Research on safety of operation and maintenance of utility tunnel based on BIM and computer technology. in Journal of Physics: Conference Series. 2021. IOP Publishing.

48.	Wu, C.-M., et al. Development and application scenario of municipal utility tunnel facility management based on BIM, 3D printer and IoT. in 2020 IEEE Eurasia Conference on IoT, Communication and Engineering (ECICE). 2020. IEEE.

49.	Yin, X., et al., A BIM-based framework for operation and maintenance of utility tunnels. Tunnelling and Underground Space Technology, 2020. 97: p. 103252.

50.	Aziz, Z., Z. Riaz, and M. Arslan, Leveraging BIM and Big Data to deliver well maintained highways. Facilities, 2017. 35(13/14): p. 818-832.

51.	Alvarez, A.P., et al., Opportunities in airport pavement management: Integration of BIM, the IoT and DLT. Journal of Air Transport Management, 2021. 90: p. 101941.

52.	Oreg, S., Resistance to change: developing an individual differences measure. J Appl Psychol, 2003. 88(4): p. 680-93.

53.	Blomqvist, I., H. Niemi, and T. Ruuskanen, Adult Education Survey 1995: Participation in Adult Education and Training in Finland. 1998: Tilastokeskus.

54.	Vass, S. and T.K. Gustavsson. The perceived business value of BIM. in eWork and eBusiness in Architecture, Engineering and Construction-Proceedings of the 10th European Conference on Product and Process Modelling, ECPPM. 2014.

55.	Bogdanova, A.N., G.A. Fedorova, and M.I. Ragulina, Forming Teachers' Awareness Of Knowledge In The Field Of Artificial Intelligence. European Proceedings of Social and Behavioural Sciences, 2021.

56.	Bespalko, V., Elements of learning process control theory. Moscow: Knowledge. Part, 1971. 2.

57.	Skatkin, M. and V. Kraevskii, Kachestvo znanii uchashchikhsia i puti ego sovershenstvovaniia (The quality of students’ knowledge and ways to improve it). Pedagogika, Moscow, 1978.

58.	Eccles, J., Expectancies, values and academic behaviors. Achievement and achievement motives, 1983.

59.	Keey, R., Risk management: An australasian view. Process Safety and Environmental Protection, 2003. 81(1): p. 31-35.

60.	Wang, M., G. Liao, and Y. Li, The relationship between environmental regulation, pollution and corporate environmental responsibility. International Journal of Environmental Research and Public Health, 2021. 18(15): p. 8018.

61.	Kadefors, A. and P. Gluch. DEVELOPMENTS IN PARTNERING: MODELS, RELATIONSHIP DURATION AND KNOWLEDGE INTEGRATION. in Proceeding from the international conference Changing Roles; New Roles, New Challenges. Noordwijk an Zee, The Netherlands, 5-9 October 2009. 2009.

62.	Alshammari, S.H., THE INFLUENCE OF TECHNICAL SUPPORT, PERCEIVED SELF-EFFICACY, AND INSTRUCTIONAL DESIGN ON STUDENTS’USE OF LEARNING MANAGEMENT SYSTEMS. Turkish Online Journal of Distance Education, 2020. 21(3): p. 112-141.

63.	Dawson, M., B. DeWalt, and S. Cleveland, The case for UBUNTU Linux operating system performance and usability for use in higher education in a virtualized environment. 2016.






\end{document}